\relax
\documentclass[letterpaper]{article} 
\usepackage{aaai22}  
\usepackage{times}  
\usepackage{helvet}  
\usepackage{courier}  
\usepackage[hyphens]{url}  
\usepackage{graphicx} 
\usepackage{natbib}  
\usepackage{caption} 
\usepackage{adjustbox}
\usepackage[algo2e]{algorithm2e}

\DeclareCaptionStyle{ruled}{labelfont=normalfont,labelsep=colon,strut=off} 
\usepackage{algorithm}
\usepackage{algorithmic}

\usepackage{newfloat}
\usepackage{listings}
\floatstyle{ruled}
\newfloat{listing}{tb}{lst}{}
\floatname{listing}{Listing}
\title{Graph-based Ensemble Machine Learning for Student Performance Prediction}
\author{
    Yinkai Wang\equalcontrib,
    Aowei Ding\equalcontrib,
    Kaiyi Guan\equalcontrib,
    Shixi Wu,
    Yuanqi Du
}
\affiliations{
    
}

\usepackage{bibentry}

\begin{document}

\maketitle

\begin{abstract}
Student performance prediction is a critical research problem to understand the students' needs, present proper learning opportunities/resources, and develop the teaching quality. However, traditional machine learning methods fail to produce stable and accurate prediction results. In this paper, we propose a graph-based ensemble machine learning method that aims to improve the stability of single machine learning methods via the consensus of multiple methods. To be specific, we leverage both supervised prediction methods and unsupervised clustering methods, build an iterative approach that propagates in a bipartite graph as well as converges to more stable and accurate prediction results. Extensive experiments demonstrate the effectiveness of our proposed method in predicting more accurate student performance. Specifically, our model outperforms the best traditional machine learning algorithms by up to 14.8\% in prediction accuracy.

\end{abstract}

\section{Introduction}


Student performance prediction, which aims to improve the teaching quality to meet the students' needs and provide proper learning opportunities and resources with identifying, extracting ,and utilizing data through different methods and techniques, has been a long-standing challenge in education~\cite{Osmanbegovic2012Data}. The deep learning models provide great performance in multiple areas, including 
Nature Language Processing\cite{Devlin2019BERTPO}, \cite{faisal2021dataset}, \cite{greff2016lstm} and graph representation learning\cite{kipf2017semisupervised}, \cite{du2021graphgt}, \cite{velivckovic2017graph}, etc. Most recent work focus on applying the rising deep learning models to the problem \cite{su2018exercise}, while rare people continue to explore the capability of traditional machine learning models. 
Many modern universities have collected large volumes of data for managing the education process. As data volumes and complexity increase, it becomes harder and harder for the university administration to handle. Advanced technologies for dealing with these data are desired yet not fully explored~\cite{Kabakchieva2012Student}. In this paper, we study the classic yet important problem to predict student academic performance with demographic, social, school-related features, where we identify three challenges in the problem, (1) the predictions are highly dependent on the students' grades rather than other demographic, social features, (2) the distributions of students' grades are imbalanced with only a few students that are below the passing scores, (3) single machine learning method fails to produce stable results across different sets of features. To this end, we propose a graph-based ensemble machine learning method that incorporates the idea of ensemble learning~\citep{Polikar2012} in the whole framework including feature selection and model prediction phases. Concerning the feature selection part, we implement an iterative algorithm that follows greedy algorithm~\citep{JIANG2018110} which filters features to reach the optimal prediction performance in every step. For the prediction model part, our method augments the predictions of multiple single machine learning methods with unsupervised clustering methods with a graph-based propagation method. In this way, it averages out the instability in a single machine learning model and constantly produces more accurate predictions. To be specific, we first select several popular machine learning prediction methods, e.g. decision tree~\citep{Song2015DecisionTM}, support vector machine (SVM)~\citep{Yongli2012Support},etc., and a list of unsupervised clustering methods,e.g. hierarchical clustering~\citep{Murtagh2012Algorithms}, DBSCAN~\cite{Schubert2017DBSCAN},k-means~\citep{durairaj2014educational}, etc. Then, we build an iterative method that dynamically ensembles results in a bipartite graph from different prediction/clustering models until convergence. Experiment results have shown that our proposed framework produces up to 14.8\% better performance than the best popular machine learning prediction methods.  Our contributions are summarized as follows:
\begin{itemize}
\item We propose a new framework that takes advantage of the idea of ensemble learning and averages out the instabilities of previous machine learning models in student performance prediction.
\item We design an iterative graph-based ensemble method that dynamically leverages and aggregates multiple supervised and unsupervised machine learning models to achieve better prediction accuracy. 
\item Extensive experiments demonstrate each part of our proposed method improves the performance of baseline models and overall further improves the prediction accuracy by up to 14.8\%.
\end{itemize}

\section{Methodology}
\label{sec:method}

\textbf{Ensemble Feature Engineering Module.} We implement an ensemble-based iterative feature selection algorithm that takes all the features from the beginning and iteratively searches for better feature sets. The idea of this algorithm implements the greedy algorithm which follows the problem-solving heuristic to make an optimal choice for every single step. Specifically, the algorithm takes measurement (i.e. importance) of each feature weighted by a list of reliable classification models. Every time, each feature is dropped from the feature set, and the classification performance (accuracy) change indicates the importance of the feature. To make the feature importance more reliable, we ensemble multiple classification models and take the best one as the current best score. Then, it drops the worst negatively-affecting feature every iteration. Additionally, to leverage the side effect of the algorithm being too greedy, we set a stop mechanism, parameterized by $k$, which stops after a continuous lower performance in $k$ iterations and rolls back to the previous best feature set. After experiments with different values of $k$, the result shows that if dropping positively-affecting feature more than $3$ times, the accuracy will keep decreasing. In practice, we set this parameter to $3$, and the list of classification models selected are [Random Forest, SVM, XGBoost]. The pseudo-code for the algorithm is shown in code block~\ref{alg:feature}.

\begin{algorithm}[h] 
\caption{Ensemble Feature Engineering Module}
\label{alg:feature}
\textbf{Data}: Student Performance Data Set

 Initialize a list of all classification models $M$, model scores $PM$, all features $G$, scores for all features $P$ as 0s, a stop counter $K$, parameterized by $k$, an empty list of temp features $T$, a best feature set $B$, a best accuracy $A$;
 
 \For{model $m$ in the model list $M$}{
    Run the model $m$;
    
    Record the prediction accuracy to $PM$;
 }
 Record best accuracy $A$ and best feature set $B$ with highest scores in $PM$;
 
 Initialize stop counter $K$=$3$;
 
 \While{condition $K$ == $0$ is not satisfied}{
 
    \For{feature $f$ in the feature list $F$}{
       Drop feature $f$;
       
       \For{model $m$ in the model list $M$}{
           Run the model $m$;
           
           Record $max(PM, prediction\:accuracy)$ to $PM$;
       }
       Add feature $f$;
    }
    \eIf{$K == 0$}{
        Resume the best feature set $B$ and best accuracy $A$;
        
        \Return the best feature set $B$, best accuracy $A$;
    }
    {\eIf{all scores in $P > 0$}{
        Drop the feature with the highest score $P$;
        
        \If{highest score $>$ best accuracy $A$}{
            Record best accuracy $A$ and best feature set $B$ with highest scores $PM$;
        }
    }{
        Drop the feature with the lowest score $P$;
        
        K $-$ = 1;
        
        Continue;
    }
    }
 }
\end{algorithm}

\begin{figure}[htbp]
    \centering
    \includegraphics[width=0.40\textwidth]{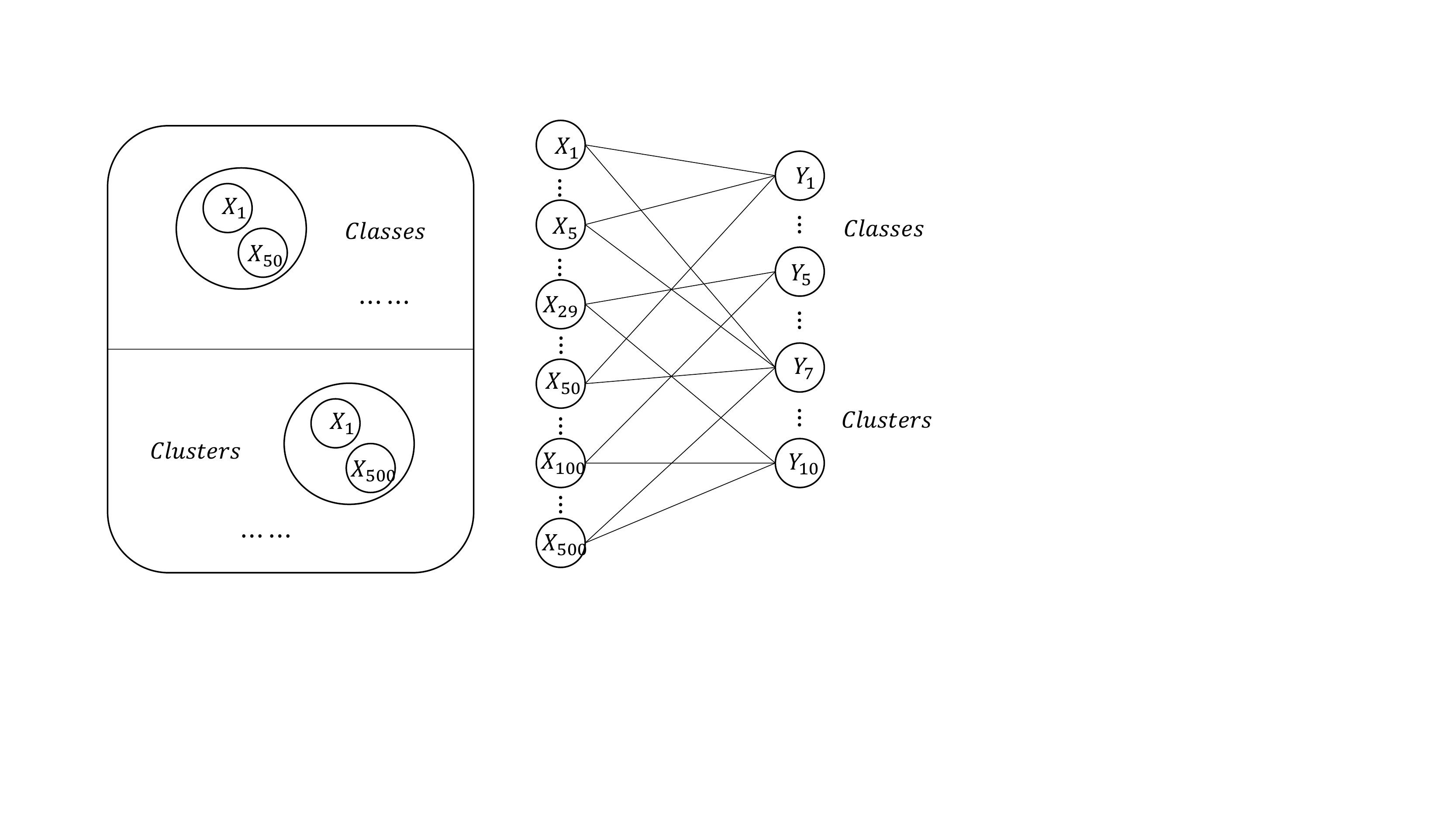}
    \caption{The groups of the graph-based ensemble classification and clusteringmodel and the bipartite graph constructed by the model.}
    \label{fig:gecm}
\end{figure}

\textbf{Ensemble Prediction Module. }We implement a graph-based ensemble classification model, which takes several reliable classifications and clustering models and aggregates the results by the message propagation over a bipartite graph to achieve better performance. The idea is to ensemble over several reliable algorithms to achieve more accurate and stable results. It is very common to ensemble the results of supervised learning models, while it is not that common to study how unsupervised learning models, such as clustering algorithms, can help in the ensemble process. Here, we aim to assist the prediction from the supervised classification learning models with uncertainty from the unsupervised clustering models. We first initialize all our models (classification, clustering) with grid-searched parameters. We construct the graph $G=(V,E)$, where nodes $V$ represent objects (either a data point $X$ or a group $Y$), and edges $E$ represent the connection between a data point and a group (class/cluster) determined by the classification/clustering algorithms. Empirically, we utilize three classification models and two clustering models. The total groups are $20$ for classification models. We leave all groups on the right side of the bipartite graphs (i.e. $40$ groups) because there is no consistent correspondence among different clustering models. Specifically, each edge is built if one classification model predicts the class of the sample or clustering algorithm clusters the samples together to one group. Because there is no supervision for the clustering algorithm, we utilize it to enhance the confidence of the classification models. We do so by calculating a confidence matrix $C \in X^{I, J}$, where $I$ is the total number of the samples and $J$ is the total number of the groups. Each entry $X_{i,j}$ represents the confidence score of a sample in a group, which is initially measured by the predicting accuracy of the classification models. The confidence score is calculated via propagation over the graph:
\begin{equation}
 X_{i,j} = X_{i,j} + \sum_{i=1}^{I}\sum_{j=1}^{J}\frac{X_{i,j}}{I \times J}
\end{equation}
Finally, the prediction class is determined by taking the label with the maximum probability, or certainty.
The groups are shown on the left and the bipartite graph is shown is the right of Fig. \ref{fig:gecm}

\section{Experiments}
\label{sec:exp}
\subsection{Experiment Set-up}
\begin{figure}[htbp]
    \centering
    \includegraphics[width=9cm]{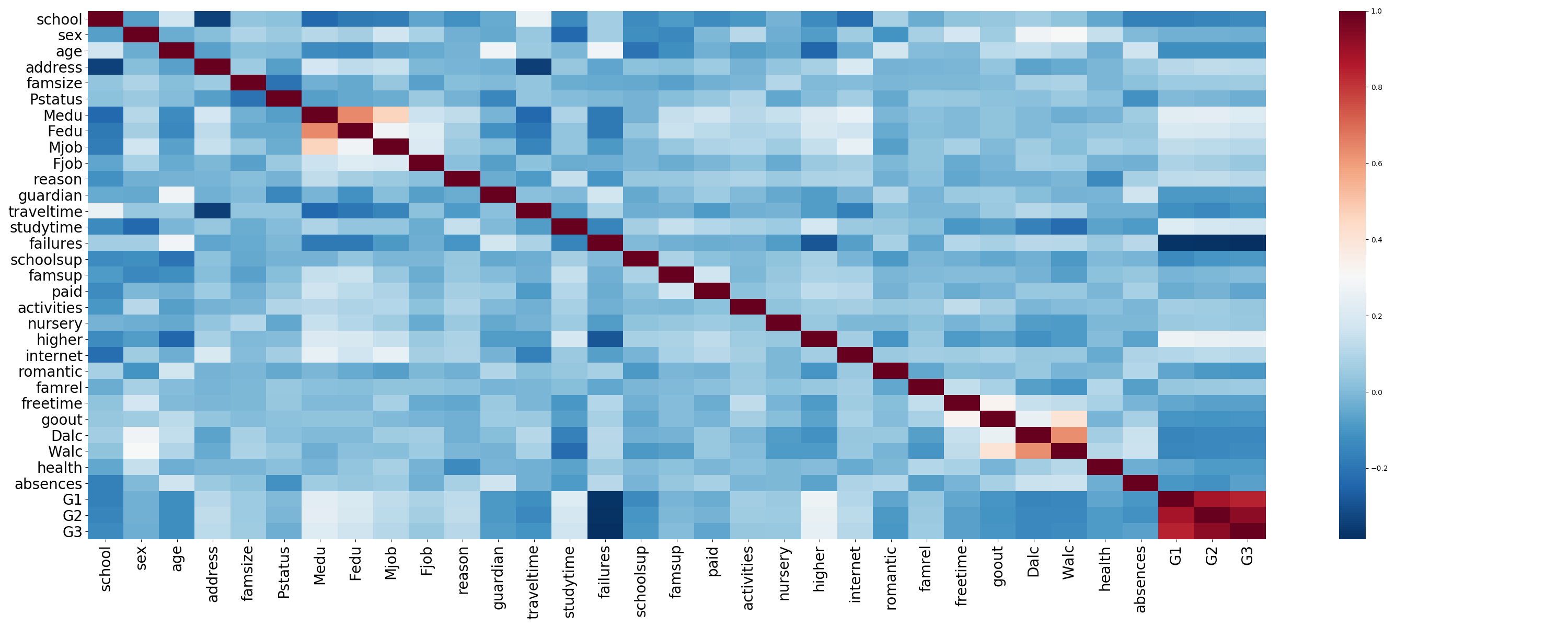}
    \caption{The heatmap shows the correlation among all the features.}
    \label{fig:heatmap}
\end{figure}

\begin{figure}[htbp]
\centering
\begin{tabular}{ccc}
\multicolumn{3}{c}{}\\
\includegraphics[width=0.3\linewidth]{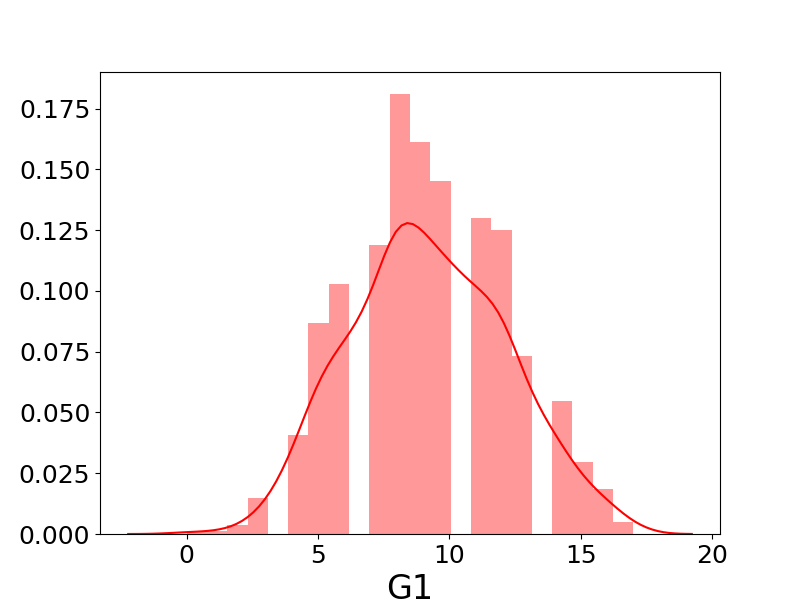} &
\includegraphics[width=0.3\linewidth]{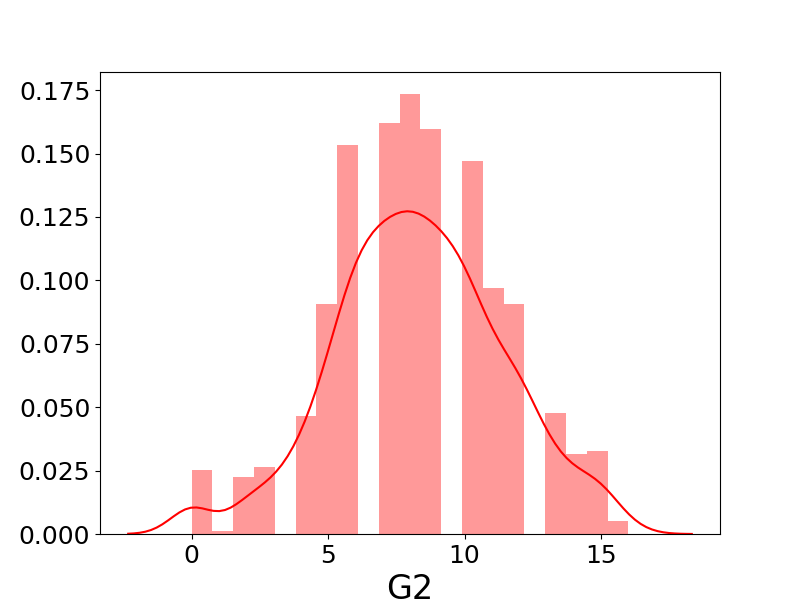} &
\includegraphics[width=0.3\linewidth]{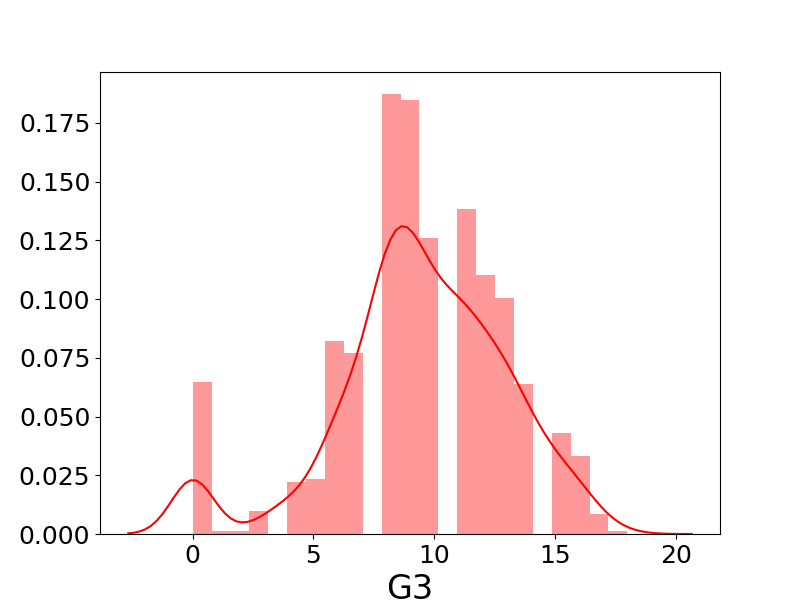} 
\end{tabular}
\caption{Score feature distributions of the dataset \cite{Dua:2019}.}

\label{fig:score_feat}
\end{figure}

\textbf{Dataset Description.} We take one commonly used benchmark dataset, Student Performance Data Set \cite{Dua:2019}, from the UCI Machine Learning Repository~\footnote{https://archive.ics.uci.edu/ml/index.php}. The dataset consists of student achievements in secondary education of two Portuguese schools. 
It contains many important factors, including student grades, demographic, social, and other school-related features. 
We analyze the dataset by visualizing the correlation among all features, in Fig.~\ref{fig:heatmap}. 
It is worth noting that the correlations among different term grades are extremely high, and the grades distributions are close to a Gaussian distribution (Fig.~\ref{fig:score_feat}.). Interestingly, we find father's education level is highly correlated with mother's education level. Mother's education level is highly correlated with the mother's job, too. However, father's education level is not correlated with father's job. This provides some statistical insights about the correlations between father/mother education levels and their jobs.
We aim to dig into other "hidden" features rather than the grade-related features in our setting.
\begin{table}[h]\small
  \centering
  \caption{Comparison of the prediction accuracy w/ and w/o resampled data.}
  \label{tab:sample}
  \begin{tabular}{|c|c|c|c|c|}
    \hline
    Models & Raw-data & Over-s & Under-s & Combined-s \\ \hline
    KNN & 0.1100 & 0.0957 & \textbf{0.1292} & 0.0861 \\\hline
    SVM & \textbf{0.1292} & 0.0287 & 0.1196 & 0.0667 \\\hline
    Random Forest & 0.1435 & \textbf{0.1531} & 0.1388 & 0.1483 \\\hline
    XGBoost & \textbf{0.1770} & 0.1579 & 0.1340 & 0.1483 \\\hline
    MLP & 0.1435 & 0.0718 & \textbf{0.1627} & 0.1005 \\\hline
\end{tabular}
\end{table}

\begin{table}[htb]
  \centering
  \caption{Comparison of the prediction accuracy w/ and w/o feature G1,G2.}
  \label{tab:comp}
  \begin{tabular}{|c|c|c|}
    \hline
    Models & w/ Grade & w/o Grade \\ \hline
    KNN & 0.2679 & 0.1100 \\ \hline
    SVM & 0.3014 & 0.1292 \\ \hline
    Random Forest & 0.4019 & 0.1435 \\ \hline
    XGBoost & 0.4498 & 0.1770 \\ \hline
    MLP & 0.2679 & 0.1435 \\\hline
    Proposed Model & \textbf{0.4641} & \textbf{0.1866} \\
    \hline
\end{tabular}
\end{table}

\begin{table}[htb]\small
  \centering
  \caption{Comparison of the prediction accuracy between the proposed model and other models.}
  \label{tab:final}
  \begin{tabular}{|c|c|c|}
    \hline
    Models & Raw-data & Feature-selected-data\\ \hline
    SVM & 0.1292 & 0.1292 \\ \hline
    Random Forest & 0.1435 & 0.1675 \\ \hline
    XGBoost & 0.1770 & 0.1627 \\ \hline
    MLP & 0.1435 & 0.1483 \\\hline
    Proposed Model & \textbf{0.1866} & \textbf{0.1922} \\
    \hline
\end{tabular}
\end{table}

\textbf{Comparison Methods Details.} We take five reliable classification models, K-nearest-neighbor, SVM, Random Forest, XGBoost and Multi-layer Perceptron (MLP) as our baselines. We take the implementation of them from the scikit-learn library\footnote{https://scikit-learn.org/}. In terms of evaluation metrics, we take accuracy scores to evaluate the prediction performance of our model. For a fair comparison, the models that we use in our ensemble system are exactly the same as the baselines, i.e. with the same parameters. We run a grid-search to find the best set of each model's parameter setting. We divide the train/validation/test set by $6/2/2$.

\begin{table*}[!h]
  \centering
  \caption{Comparison of the prediction accuracy w/ and w/o feature engineering.}
  \label{tab:pred}
  \begin{tabular}{|c|c|c|c|c|c|}
    \hline
    Models & Raw-data & PCA-90 & PCA-95 & Tree-based & Ensemble-based\\ \hline
    KNN & 0.1100 & 0.1244 & 0.1388 & \textbf{0.1579} & 0.1483 \\\hline
    SVM & 0.1292 & 0.1388 & 0.1388 & \textbf{0.1388} & 0.1292 \\\hline
    Random Forest & 0.1435 & 0.1053 & 0.1244 & 0.1627 & \textbf{0.1675}\\\hline
    XGBoost & 0.1770 & 0.0766 & 0.1100 & 0.1483 & \textbf{0.1771}\\\hline
    MLP & 0.1435 & 0.0901 & 0.1069 & 0.1738 & \textbf{0.1866}\\\hline
\end{tabular}
\end{table*}

\subsection{Experiment Results}
\textbf{Ensemble Feature Engineering Evaluation}
The final selected features by our model are school, sex, age, address, famsize, Pstatus, Fedu, Mjob, Fjob, reason, guardian, traveltime, studytime, failures, schoolsup, famsup, paid, activities, nursery, higher, internet, romantic, famrel, freetime, Dalc, Walc and health. 
We show the prediction accuracy of all the baseline models on data provided by different feature selection algorithms in Table.~\ref{tab:pred}. Clearly, our ensemble feature selection model outperforms the raw data in all baseline models. To compare with other feature selection techniques, such as Principle Component Analysis (PCA) \cite{wold1987principal}, and tree-based feature selection algorithm. For most of the cases, our model achieves the best performance. While the tree-based approach is also competitive while pairing with KNN or SVM. We suspect the reason is that our model is a little dominated by the best classification models during the feature selection phase. KNN and SVM are normally the worst models across all the models. Therefore, our selected features are not perfectly aligned with these two models. Besides higher accuracy, our model provides more explainability for researchers with a stronger feature selection module.

\textbf{Data Sampling Evaluation}
In order to see whether popular data re-sampling techniques are suitable for our problem, we select three well-known data re-sampling algorithms and take the implementation from the imblearn library~\footnote{https://imbalanced-learn.org/stable/}. To be specific, we take one oversampling, one undersampling and one combined-sampling algorithm, respectively. Specifically, we utilize SMOTE \cite{chawla2002smote} for oversampling, TomekLinks for undersampling and SMOTETomek for combined sampling \cite{more2016survey}. The results are shown in table \ref{tab:sample}. Even though the feature re-sampling techniques show some promises in some cases, it is still not statistically reliable for us to incorporate into our system.

\textbf{Ensemble Prediction Evaluation}
We study the effectiveness of our ensemble prediction model by comparing it with other baseline models mentioned above. The results in Table. \ref{tab:comp} suggest that our model is the best model, either with the presence of the grade features, or with the absence of the grade features. It is also worth noting that XGBoost ranks second in both settings, which shows the great power of the algorithm. Random forest also achieves stable performance compared to KNN, SVM and MLP. Finally, KNN performs the worst in our testing. Additionally, table \ref{tab:comp} suggests that our model is more stable as it achieves the best performance in both settings.

\textbf{Ablation Study}
In table \ref{tab:final}, we show the ablation study of our system, in which we take out one component and see whether the other component improves the performance compared to the baseline models. Clearly, both the feature engineering module and ensemble prediction module improve the result of our system.

\section{Conclusion}
\label{sec:conclusion}
In this paper, we propose a graph-based ensemble machine learning method for student academic performance prediction, which consists of ensemble feature engineering and ensemble prediction modules. The extensive experiments have shown that each component of our system outperforms any single machine learning method. Overall, the system further improves the prediction accuracy by $14.8$\%. In the future, we plan to investigate how this method works in other large-scale datasets.

\bibliography{sigproc}

\end{document}